\documentclass[letterpaper, 10 pt, conference]{ieeeconf}  

\IEEEoverridecommandlockouts    
\let\labelindent\relax          

\usepackage{graphicx}    
\usepackage{subcaption}  
\usepackage{mathptmx}    
\usepackage{times}       
\usepackage{hyperref}    
\usepackage{tikz}        
\usepackage{ulem}        
\usepackage{amsmath}     
\usepackage{amssymb}     
\usepackage{bm}          
\usepackage{siunitx}     
\usepackage{mathtools}   

\usepackage[utf8]{inputenc}  
\newcommand\scaleddot{\scalebox{.89}{.}}
\makeatletter
\renewcommand{\dddot}[1]{%
  {\mathop{\kern\z@#1}\limits^{\makebox[0pt][c]{\vbox to-2.2\ex@{\kern-\tw@\ex@
  \hbox{\normalfont\scaleddot\kern-0.5pt\scaleddot\kern-0.5pt\scaleddot}\vss}}}}}
\renewcommand{\ddddot}[1]{%
  {\mathop{\kern\z@#1}\limits^{\makebox[0pt][c]{\vbox to-2.2\ex@{\kern-\tw@\ex@
  \hbox{\normalfont\scaleddot\kern-0.5pt\scaleddot\kern-0.5pt\scaleddot\kern-0.5pt\scaleddot}\vss}}}}}
\makeatother

\renewcommand{\vec}[1]{\mathbf{#1}}

\newcommand\ballPosScalar{b}
\newcommand\ballPos{\vec{\ballPosScalar}}

\newcommand\ballVel{\dot{\ballPos}}

\newcommand\ballAcc{\ddot{\ballPos}}
\newcommand\eePosScalar{x}
\newcommand\eePos{\vec{\eePosScalar}}

\newcommand\eeVel{\dot{\eePos}}

\newcommand\eeAcc{\ddot{\eePos}}

\newcommand\handNormal{\vec{e}_h}
\newcommand\gravity{\vec{g}}

\newcommand\flightTime{T_\mathrm{flight}}

\newcommand\cycleTime{T_\mathrm{cycle}}

\newcommand\DES{\mathrm{des}}
\newcommand\TD{\mathrm{TD}}
\newcommand\TO{\mathrm{TO}}
\newcommand\PostTO{\mathrm{postTO}}
\newcommand\PreTD{\mathrm{preTD}}
\newcommand\ContactPrev{\mathrm{cont\_prev}}

\newcommand{\stkout}[1]{\ifmmode\text{\sout{\ensuremath{#1}}}\else\sout{#1}\fi}

\title{\LARGE \bf Beyond the Cascade: Juggling Vanilla Siteswap Patterns}

\author{\vspace{-10pt}Mario Gomez Andreu$^{1}$, Kai Ploeger$^{1}$ and Jan Peters$^{1,2,3}$
\thanks{$^{1}$ The authors are with the Technical University of Darmstadt, Germany}
\thanks{$^{2}$ Jan Peters is with the German Research Center for AI (DFKI)}
\thanks{$^{2}$ Jan Peters is with the Hessian Centre for Artificial Intelligence}
}%

\begin{document}

\maketitle
\thispagestyle{empty}
\pagestyle{empty}

\begin{abstract}

Being widespread in human motor behavior, dynamic movements demonstrate higher efficiency and greater capacity to address a broader range of skill domains compared to their quasi-static counterparts.
Among the frequently studied dynamic manipulation problems, robotic juggling tasks stand out due to their inherent ability to scale their difficulty levels to arbitrary extents, making them an excellent subject for investigation.
In this study, we explore juggling patterns with mixed throw heights, following the vanilla siteswap juggling notation, which jugglers widely adopted to describe toss juggling patterns.
This requires extending our previous analysis of the simpler cascade juggling task by a throw-height sequence planner and
further constraints on the end effector trajectory. These are not necessary for cascade patterns but are vital to achieving patterns with mixed throw heights.
Using a simulated environment, we demonstrate successful juggling of most common 3-9 ball siteswap patterns up to 9 ball height, transitions between these patterns, and random sequences covering all possible vanilla siteswap patterns with throws between 2 and 9 ball height.
\href{https://kai-ploeger.com/beyond-cascades}{https://kai-ploeger.com/beyond-cascades}
\end{abstract}

\section{INTRODUCTION}

Dynamic manipulation strategies frequently demand high speeds from the manipulator in use, potentially impeding its dexterity.
%
However, they hold promise in addressing challenges like torque constraints and under-actuation, thus streamlining hardware requirements to a considerable extent.
%
Juggling distinguishes itself within the domain of dynamic manipulation tasks, requiring the simultaneous combination of high speed and intricate dexterity.
%
Toss juggling inherently presents the challenge of under-actuation, given its usual scenario of controlling more objects than available hands.
%
This requires careful planning of object-hand interactions.
%
In our previous work~\cite{ploeger2022controlling}, we investigated the processes of clean contact switches at the moments of touch-down and take-off in the context of uniform cascade juggling.
%
However, as we extend the juggling task in this study to include mixed throws of greatly varying heights, it becomes apparent that further constraints during the carry and dwell phases between contact switches are necessary to maintain desired contacts and avoid undesired contacts.
In this work, we complete the set of time-continuous movement constraints required for planning general toss juggling trajectories.
%
Employing a high-level planner for seamless siteswap navigation, we successfully execute all possible vanilla siteswap juggling patterns with throw heights between 2 and 9 in simulation scenarios.
Comparing to~\cite{ploeger2022controlling} as a baseline,
we show the necessity and the sufficiency of the proposed constraint set.   

\begin{figure}
    \centering
    \vspace{5pt} 
    \includegraphics[width=\columnwidth, trim={6.6cm 2.6cm 6.6cm 0.6cm}, clip]{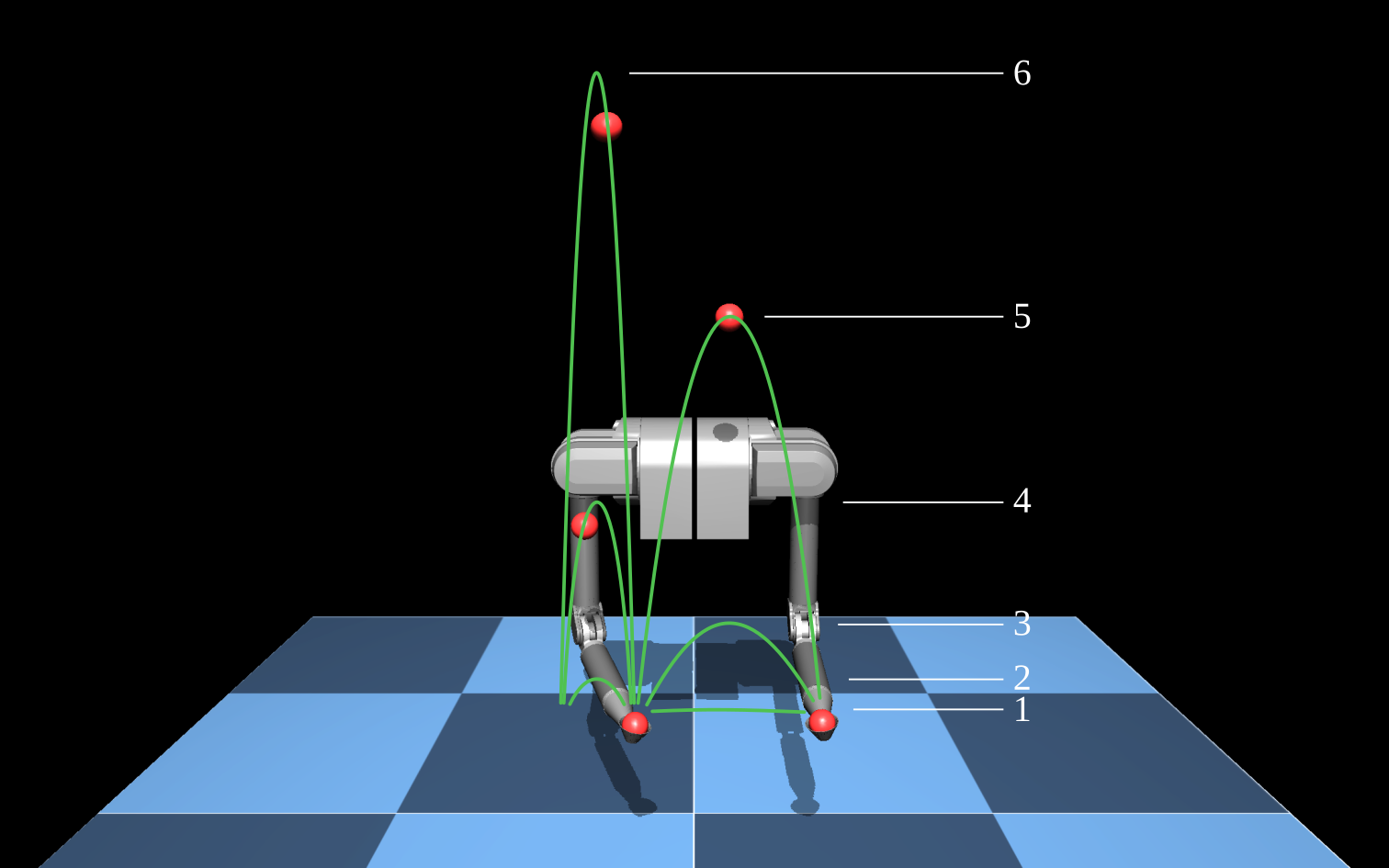} 
    \vspace{-12pt}
    \caption{At a given throw frequency, throw heights are restricted to a discrete set. Successful juggling patterns of up to height 9 reach $4.53\si{m}$ tall, as measured from catch height, and include up to $9$ balls simultaneously.}
    \label{fig:throw_heights}
    \vspace{-0.7cm}
\end{figure}

\vspace{-3pt}
\subsection{Problem Statement}
\vspace{-2pt}

Our primary aim is to attain proficiency in juggling all two-handed juggling patterns that adhere to the assumptions of the siteswap notation.
These assumptions include:
a)~throws occurring at discrete and equidistant points in time called beats,
b)~throws being alternated between distinct hands,
c)~hands holding at most a single ball at a time, and
d)~balls being held in hand for a non-zero constant dwell duration.
%
Patterns conforming to these assumptions provide the option to select from a discrete set of feasible throw heights for each individual throw, as illustrated in Figure~\ref{fig:throw_heights}.
When a ball is thrown to height $N$, it remains airborne for a duration, allowing it to be thrown again after exactly $N$ beats.
Equivalently an $N$-ball uniform pattern consists of only throws of height $N$, which we will refer to as $N$-throws.
We consider throws up to height 9, which, in our case, is equivalent to $4.53\si{m}$.
A juggling pattern is characterized by a recurring sequence of throws,
each potentially differing in height.
Precise planning of patterns and transitions is essential
to avoid scenarios in which one hand is required to catch and throw multiple balls simultaneously.

\vspace{2pt}
\textbf{Contributions:}
We extend the previously established set of trajectory constraints dealing with contact switches by additional constraints for contact management between contact switches
%
and demonstrate the capability to successfully juggle any number of balls in any pattern with throw heights between $2$ and $9$ utilizing this complete set of constraints.

\subsection{Related Work on Robot Juggling}
\label{sec:robot_juggling}

Early investigations of juggling machines involved equipping one-degree-of-freedom open-loop automata with funnel-shaped end effectors,
%
allowing for limited open-loop stability by dissipating kinetic energy and reducing variance in the ball position at touchdown,
%
including Claude Shannon's juggling automata~\cite{atkeson2017shannon, schaal1993open}, with three and five-ball lift bounce cascades,
%
a subsequent regular three-ball cascade~\cite{machines1002007juggling} automaton,
%
and a five-ball cascade through linear actuation~\cite{burget2010visual}.
%
Data-driven approaches optimized open-loop movement primitives for two-ball one-handed juggling through trial and error.
%
In~\cite{sakaguchi1991study}, the movement primitive was updated in a model-based fashion, accounting for ballistics,
and in one of our previous works~\cite{ploeger2021high}, the movement primitive was updated in a black-box fashion, achieving close to two hours of sustained juggling.

Including feedback on the ball's states has shown to be challenging.
For instance, camera systems in~\cite{burget2010visual} and \cite{ploeger2021high} were used solely to detect ball drops.
%
In~\cite{kober2012playing}, a combination of a hand-tuned throwing movement and a learned catching movement conditioned on the ball state resulted in 3-ball human-robot partner juggling.
%
While all previous approaches used funnel-shaped hands, \cite{kizaki2012two} utilized a three-fingered gripper to juggle two balls in one hand,
combining kinematically planned catching movements with a hand-tuned movement primitive for throwing.
%
In our previous work~\cite{ploeger2022controlling}, we explored which constraints have to be fulfilled by the robot hand for cascade juggling.
We showed that the maximum number of balls that can be juggled in a cascade is not limited by the throw height, but rather by the distance between the hands.
Using a simulated environment, we achieved a stable 17-ball cascade, which represents the theoretical maximum for the chosen setup.

Other related tasks include paddle juggling, where the robot keeps a ball bouncing on an actuated paddle. 
%
Previous works solved the planar one-ball~\cite{buhler1990stable} and two-ball~\cite{buhler1991planning} cases closed-loop,
%
extended to the spatial case~\cite{rizzi1993further},
%
and showed open-loop stability in paddle juggling can be achieved by slowing down the paddle while hitting the ball~\cite{schaal1993open}.
%
While most approaches focus on uniformly throwing or batting balls to approximately the same height,
\cite{buhler1990stable} showed the existence of multi-cycle stable loops in paddle juggling.
%
In this work, we will systematically investigate the case of toss juggling with throws to varying heights.
This type of toss juggling is typically referred to as siteswap juggling.

\begin{figure}[h]
    \centering   
    \includegraphics[width=0.8\columnwidth]{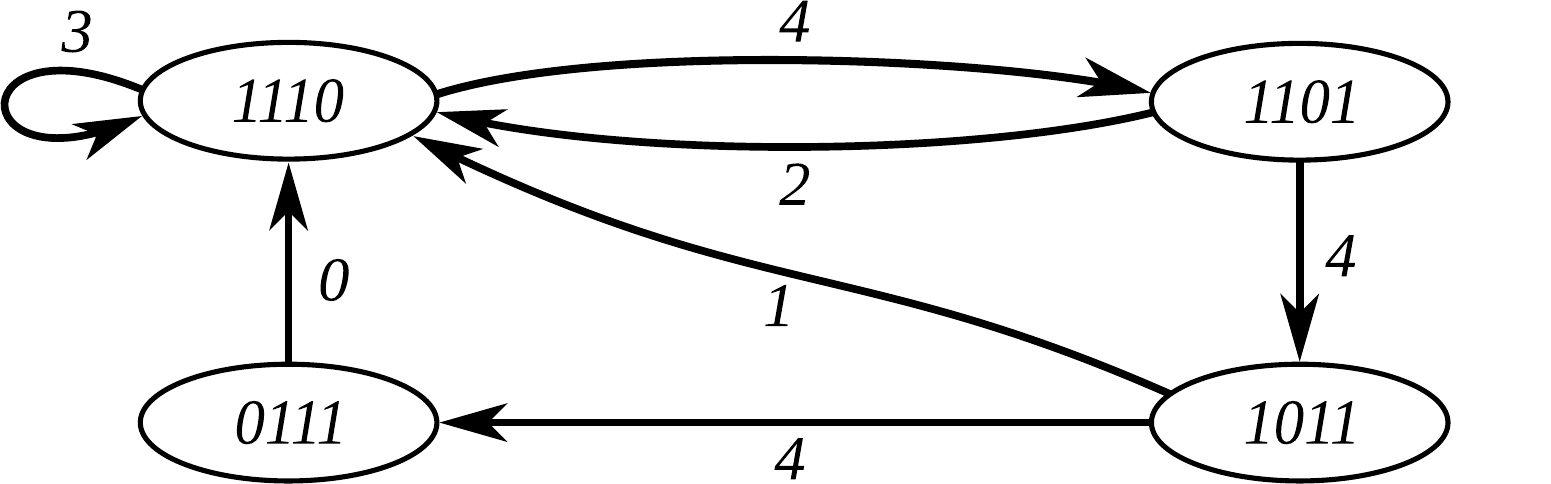}
    \caption{The three ball siteswap graph of height 4: Juggling sequences are planned on this type of graph. Each transition represents a throw of specified height and each loop in the graph is a unique pattern.}
    \label{fig:3_ball_graph}
    \vspace{-0.5cm}
\end{figure}

\subsection{Introduction to Siteswap Juggling}
\label{sec:siteswap_juggling}


While in the cascade and fountain patterns all objects are thrown to the same height,
%
siteswap patterns allow for throws of mixed heights.
%
Throw heights are not represented in meters but in the number of equidistant throw beats until they are thrown again,
resulting in a finite set of available throw heights,
as shown in Figure~\ref{fig:throw_heights}.
A ball thrown to height $3$ will stay airborne long enough to be caught and thrown again after $3$ beats.
%
We will refer to the action of throwing a ball to height $3$ as a $3$-throw or as throwing a $3$.
Analogously, a $3$-catch or catching a $3$ refers to the action of catching a ball that was thrown to height $3$.
%
The exclusive usage of $N$-throws results in the $N$-ball uniform cascade or fountain pattern.
%
To mix throw heights, we can define a juggling sequence such as $(4,2,3,4,2,3,4\dots)$,
%
which for notational convenience is typically written as $423$, assuming that it repeats endlessly.
%
Not all number sequences are valid juggling sequences. For example, an $N$-throw can not immediately be followed by an $N$-1-throw since both balls would have to be caught simultaneously by the same hand.
%
The number of required objects is given by the average of all throw heights in the sequence.
We can see that the $423$ pattern requires three balls, while the $633$ is a four-ball pattern.
%
Non-vanilla siteswap notations have extended to synchronous throws, multi-ball throws and catches, as well as partner juggling with arbitrary numbers of hands, as detailed in~\cite{polster2003mathematics}.
%
In this work, we will focus on strict vanilla siteswaps, treating the case of two-handed asynchronous juggling to a constant beat.


An abstract juggling state can be defined as an N-hot vector, denoting at which future beats an object is scheduled to be caught and thrown.
%
The state ${}^{(3)}s_B = \left[1, 1, 1, 0\right]^\intercal$ is commonly referred to as the three-object ground state,
denoting that all objects will be caught and thrown in the following three beats.
%
Any other state is called an excited state.
%
Throwing a $4$ from ${}^{(3)}s_B$ results in the state $\left[1, 1, 0, 1\right]^\intercal$, indicating an empty hand on the third beat.
From here, the ground state can be reached through a $2$-throw, while other throws transition to different excited states. 
%
More formally, an $i$-throw transition shifts the state vector forward by one position and sets the $i$-th entry to $1$.
%
If the $i$-th entry of the shifted state is already set to $1$, the transition is not allowed,
to avoid having to catch multiple balls at the same time in the future.
%
A leading $0$ means no ball is currently held in the throwing hand, resulting in a single beat idle time denoted as $0$-throw.
%
Given a chosen maximum throw height, these definitions of states and transitions allow the construction of a directed siteswap graph as shown in Figure~\ref{fig:3_ball_graph}.
%
Each valid siteswap pattern is represented by a loop in the graph.

%
Low throw numbers can potentially turn infeasible depending on the number of hands and the dwell ratio $r$,
%
which is defined as the fraction of the cycle time in which a hand holds a ball,
%
Considering two hands, a $1$-throw becomes infeasible at a dwell ratio $r\geq0.5$,
as it would require a flight time $\flightTime\leq0$,
unless holding the object with both hands simultaneously is allowed.
%
Most humans prefer to juggle close to $r\approx0.7$ but achieve $1$-throws by timing variations.

\section{Integrated Ball and Hand Trajectory Generation for Vanilla Siteswap Juggling}

We model the trajectory planning for vanilla siteswap juggling as a bi-level hierarchical problem.
%
First, future ball trajectories are planned.
These are fully defined by when and where hand-ball contact switches occur.
%
Subsequently, robot trajectories are planned to realize these contact switches.
%
The ball trajectory planning~\ref{sec:ball_trajectory_planning} draws from common knowledge in the juggling community~\cite{polster2003mathematics}
and Section~\ref{sec:hand_trajectory_planning} reintroduces previous work~\cite{ploeger2022controlling}.
The methodological contributions of this work are the novel constraints for continuous contact management required to transition from uniform juggling to mixed throw heights
in Sections~\ref{sec:premature_contacts} and ~\ref{sec:contact_management}

\subsection{Ball Trajectory Planning for Vanilla Siteswap Patterns}
\label{sec:ball_trajectory_planning}

Planning of possible future ball trajectories comes down to navigating the siteswap graph defined in Section~\ref{sec:siteswap_juggling}.
%
Each siteswap pattern corresponds to a loop in the graph.
%
Therefore, juggling a pattern breaks down into finding the corresponding loop in the directed graph and subsequently finding a path from the current state to the loop.
%
We always start from the ground state representing a uniform cascade or fountain pattern and generate the shortest path from the ground state to a node in the target pattern loop using Dijkstra's algorithm.
%
Similarly, when transitioning between two different patterns, we employ Dijkstra's algorithm to generate a path from a state within the current pattern to a state within the target pattern.
%
If the sets of traversed states for the current and target patterns overlap, a trivial transition sequence of zero length exists.

Given a chosen constant cycle time $\cycleTime$ and dwell ratio $r$,
the throw number $a_i$ of the $i$-th throw fully defines the corresponding flight time
${\flightTime}_i=(a_i-2r)\frac{\cycleTime}{2}$.
%
We define the desired cartesian takeoff $\ballPos_{\TO, \DES}$ and touchdown position $\ballPos_{\TD, \DES}$ for each hand as constant parameters for all throws.
%
Assuming constant gravitational acceleration $\gravity$ and zero air drag,
the resulting required takeoff velocities
\begin{equation}
\ballVel_{\TO, \DES_i} = (\ballPos_{\TD, \DES} - \ballPos_{\TO, \DES} - 0.5 \gravity {\flightTime}_i^2) / {\flightTime}_i
\end{equation}
fully define the desired contact switches.

\subsection{Basic Catch-and-Throw Robot Trajectory Optimization}
\label{sec:hand_trajectory_planning}

The hand trajectories need to pass through the given contact switches, make sure to avoid all ball contacts during the vacant time, and ensure constant ball contact during the dwell time.
%
All planned trajectories encompass one hand cycle, from one takeoff to the subsequent takeoff, as visualized in Figure~\ref{fig:constraintcycle}.
%
Building on~\cite{ploeger2022controlling},
%
we find a sequence of piecewise constant jerks in joint space,
that minimizes the integral of squared joint accelerations,
through direct single shooting
utilizing the Pinocchio~\cite{carpentier2019pinocchio}, CasADi~\cite{andersson2019casadi}, and IPOPT~\cite{wachter2006implementation} libraries.
%
The planned trajectory must fulfill a set of task space constraints, shaping the movement.
Through a forward kinematics model in the optimization loop,
the set of task space constraints defines a manifold of feasible trajectories in the joint space of the robot.
%
By applying an equidistant time discretization, we introduce an approximation to the planning problem since the resulting trajectories are effectively only constrained to a piece-wise tangent space of the time-continuous constraint manifold.
While a denser time discretization provides a more accurate approximation and a potentially better trajectory, a course time discretization reduces the number of decision variables and, indirectly, the solver time, which is crucial for real-time applicability.

\definecolor{primarycolor}{HTML}{305090}     
\definecolor{quaternarycolor}{HTML}{608FBF}  
\definecolor{secondarycolor}{HTML}{AA0000}   
\definecolor{tertiarycolor}{HTML}{60C760}    
\definecolor{tertiarycolorfade}{HTML}{C0F0C0}    
\definecolor{quinarycolor}{HTML}{ffffff}     

    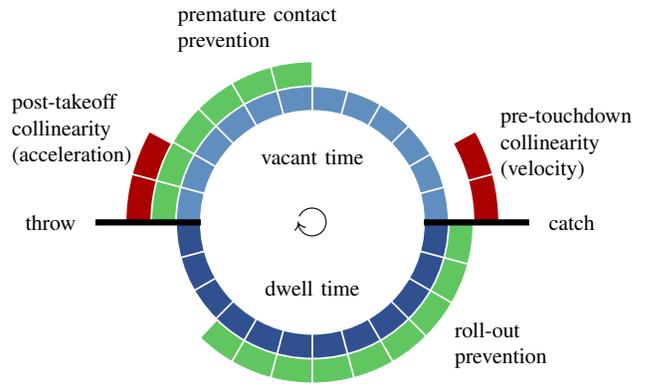
\begin{figure}[t]
        \centering
            \resizebox{\columnwidth}{!}{
            \begin{tikzpicture}
                \begin{scope}[line width=4mm,rotate=0]
                    \draw[color=quaternarycolor] (0:2cm) arc (0:180:2cm);  
                    \draw[color=primarycolor]    (180:2cm) arc (180:360:2cm); 
                    \draw[color=secondarycolor]  (151:28mm) arc (151:180:28mm); 
                    \draw[color=secondarycolor]  (0:28mm) arc (0:29:28mm);  
                    \draw[color=tertiarycolor]   (90:24mm) arc (90:180:24mm); 
                    \draw[color=tertiarycolor]   (0:24mm) arc (0:-134:24mm);  
        
                    \newcount\mycount
                    \foreach \angle in {0,150,...,3599}
                    {
                        \mycount=\angle\relax
                        \divide\mycount by 10\relax
                        \draw[quinarycolor, thick] (\the\mycount:18mm) -- (\the\mycount:32mm);
                    }
        
                    \draw (0,0)  node {\rotatebox{90}{\scalebox{2}{$\circlearrowright$}}};
                    \draw (90:1.5cm) node[below] {vacant time};
                    \draw (270:1.5cm) node[above] {dwell time};
                    \draw (-45:28mm) node[right,align=left] {roll-out\\prevention };
                    \draw (110:26mm) node[above,align=left] {premature contact\\prevention};
                    \draw (25:30mm) node[right,align=left] {pre-touchdown\\collinearity\\(velocity)};
                    \draw (150:30mm) node[left,align=left] {post-takeoff\\collinearity\\(acceleration)};
                    \draw (180:35mm) node[left,align=left] {throw};
                    \draw (0:35mm) node[right,align=left] {catch};
        
                    \draw[quinarycolor, thick] (0,0) circle (2.2cm) circle (1.8cm) circle (2.6);
                
                    \draw[black,line width=1mm] (0:18mm) -- (0:35mm);
                    \draw[black,line width=1mm] (180:18mm) -- (180:35mm);
                \end{scope}
                
                \node[text width=1.5cm] at (0,0) {};
            \end{tikzpicture}
        }
        \vspace{-25pt}
        \caption{Every planned trajectory starts and ends at takeoff (throw) and is discretized into 24 time steps.
                 During the dwell time, a ball rests within the hand. Hands move toward the incoming ball during the vacant time. 
                 Post-takeoff and pre-touchdown, ball and hand movements must be collinear to ensure clean contact switches~(\ref{sec:hand_trajectory_planning}).
                 During the vacant time, premature contacts with low incoming balls need to be avoided~(\ref{sec:premature_contacts}),
                 and during dwell time balls need to be kept from rolling out of the hand~(\ref{sec:contact_management}).}
        \label{fig:constraintcycle}
        \vspace{-10pt}
    \end{figure}

\begin{figure}[h]
    \centering   
    \includegraphics[width=0.7\columnwidth]{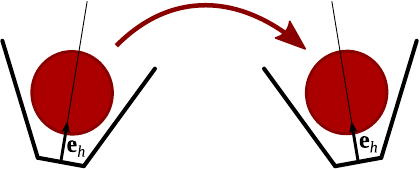}
    \caption{To achieve clean contact switches balls are kept close to the symmetry line $\handNormal$ of the funnel-shaped hand after takeoff (throw: left) and prior to touchdown (catch: right).}
    \label{fig:throwing_and_catching}
\end{figure}

%
To catch a ball at a predicted touchdown time  $\tilde{t}_\TD$, and position $\tilde{\ballPos}_\TD$, it needs to be intercepted by the hand movement
\begin{equation}
     \eePos(\tilde{t}_\TD) = \tilde{\ballPos}_\TD.
    \label{eq:catching_position}
\end{equation}
%
Velocity matching can be advantageous in preventing bouncing but is not necessary given sufficiently damped setups.
%
%
Throwing requires slightly more consideration.
%
To ensure that the ball reaches its target, the hand position and velocity
\begin{align}
    \eePos(t_\TO) &= \ballPos_{\TO, \DES} \label{eq:throw_constr_pos}, \\
    \eeVel(t_\TO) &= \ballVel_{\TO, \DES} \label{eq:throw_constr_vel}
\end{align}
has to match the previously defined contact switches at takeoff time $t_\TO$,
which is the end of the planned trajectory.
%
To break contact at the desired time, the hand acceleration
\begin{align}
    \eeAcc(t_\TO) = \gravity \label{eq:throw_constr_acc}
\end{align}
needs to match the gravitational pull on the ball at $t_\TO$.
%
%
Lateral hand movements right after takeoff can reestablish contact with the ball, leading to an unintended redirect of the ball.
%
For the case of our cone-shaped hand, we avoid these unintended collisions by keeping the ball close to the symmetry axis $\handNormal$ of the hand for a short duration $T_\PostTO$ after takeoff, as shown in Figure~\ref{fig:throwing_and_catching}.
%
Due to constraints~(\ref{eq:throw_constr_pos}-\ref{eq:throw_constr_acc}),
this can be enforced through the \emph{post-takeoff constraint}
\begin{equation}
    \forall t \in (t_\TO, t_\TO +  T_\PostTO]:\ \vec{0} = (\eeAcc - \ballAcc) \times \handNormal,
    \label{eq:takeoff_acceleration} \\
\end{equation}
restricting the relative acceleration between the hand and the ball to the direction of the hand normal $\handNormal$.
%
In the case of zero air drag, we assume the airborne ball to accelerate at a constant gravitational pull $\gravity$ and substitute $\ballAcc(t) \equiv \gravity$.
%
%
During touchdown, the ball must be approached from a feasible direction to prevent it from bouncing off the outside or edge of the hand.
%
Analogous to~(\ref{eq:takeoff_acceleration}), for a short duration $T_\PreTD$ before touchdown, the ball can be kept close to the symmetry axis of the hand by the \emph{pre-touchdown constraint}
\vspace{-1pt}
\begin{align}
    \forall t \in [\tilde{t}_\TD - T_\PreTD, \tilde{t}_\TD):\ & \vec{0} = (\eeVel - \ballVel) \times \handNormal, \label{eq:td_collinearity}
\end{align}
\vspace{-1pt}
restricting the relative velocity between the hand and the ball to the direction of the hand normal $\handNormal$.
%
In previous work~\cite{ploeger2022controlling}, this constraint has shown to be challenging to optimize numerically, motivating the approximation
\vspace{-1pt}
\begin{align}
    \forall t \in [\tilde{t}_\TD - T_\PreTD, \tilde{t}_\TD):\ & \vec{0} = \eeVel \times \tilde{\ballVel}  \label{eq:td_collinearity_approx}
\end{align}
\vspace{-1pt}
as a more tractable alternative that we will substitute with.
%
This approximation holds sufficiently well, as long as the hand is roughly directed toward the ball.
which is the case for incoming balls of height 4 and upward
in a human-like two-handed setup.

\subsection{Premature Ball-Hand Contacts}
\label{sec:premature_contacts}
%
In siteswap juggling, a unique problem arises. 
A follow-through hand movement releasing a high throw may intersect the trajectory of an incoming ball thrown at a low height.
%
We circumvent this issue by scheduling a minimum distance $d_\ContactPrev(t)$ between the hand and the incoming ball during the dwell time in a novel \emph{premature contact avoidance constraint} 
\vspace{-1pt}
\begin{equation}
    \forall t \in [t_\TO, t_\TD]: \lVert \eePos -  \ballPos \rVert > d_\ContactPrev(t). \label{eq:ball_avoidance_simple}
\end{equation}
\vspace{-1pt}
%
In the extreme case of an outgoing 9-throw followed by an incoming 3-throw, constraint~(\ref{eq:ball_avoidance_simple}) gives rise to the undesired local optimum of the hand moving over and around the incoming ball instead of under.
%
To avoid this option, we schedule the horizontal and vertical hand-ball displacements
\vspace{-15pt}
\begin{align}
    \forall t \in [t_\TO, t_\TD]:\ & \eePosScalar_z - \ballPosScalar_z < d_\mathrm{vert}(t), \label{eq:td3_vertical} \\
    \forall t \in [t_\TO, t_\TD]:\ & \lVert \eePos_{xy} - \ballPos_{xy} \rVert < d_\mathrm{horz}(t) \label{eq:td3_horizontal}
\end{align}
for incoming balls of height 3 during the vacant time.
%
This also addresses the problem of the \emph{pre-touchdown constraint} approximation~(\ref{eq:td_collinearity_approx}) not holding for $3$-throws.

\subsection{Ball Roll-Out during Carry Phase}
\label{sec:contact_management}
\begin{figure}
    \centering
    \hspace{0.5cm}
    \subcaptionbox{Frictionless Orbit\label{fig:spiral}}{\includegraphics[width=0.33\columnwidth]{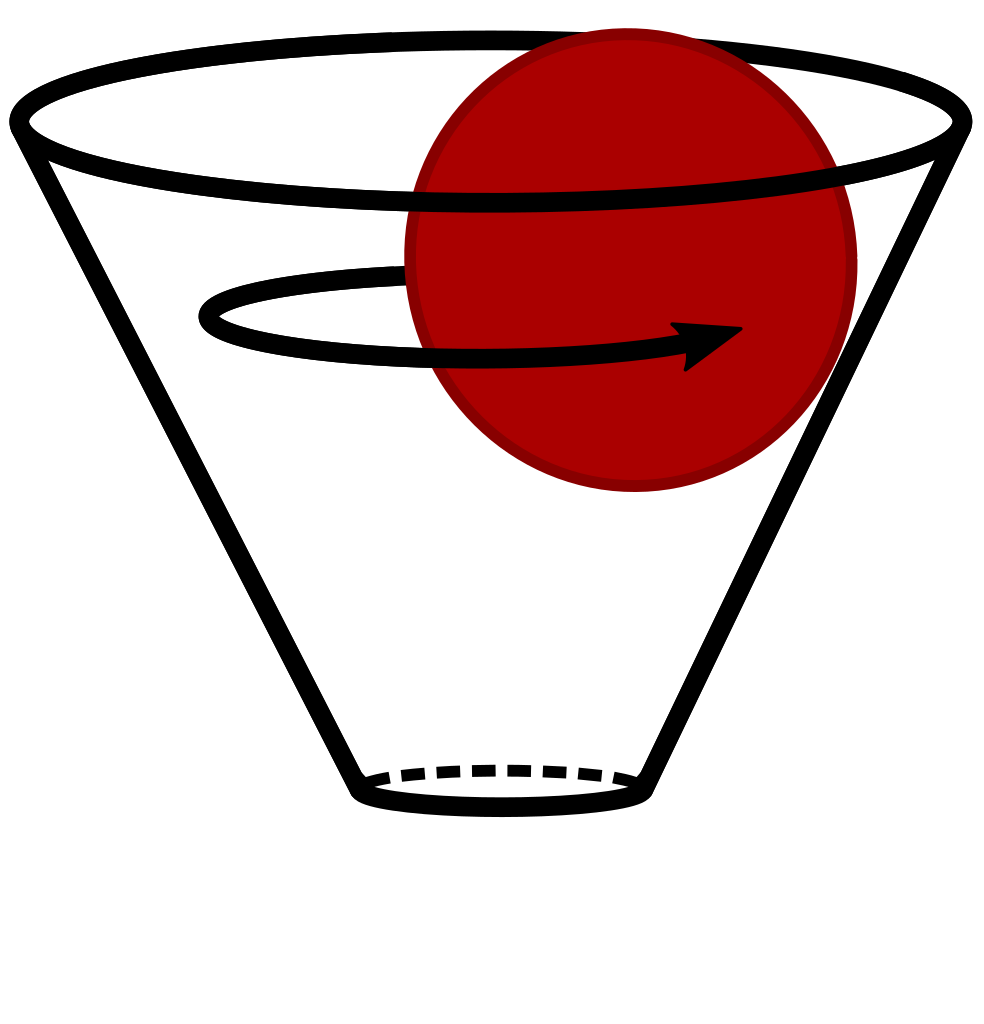}}
    \hfill
    \subcaptionbox{Roll-Out Constraint\label{fig:rollout}}{\includegraphics[width=0.45\columnwidth]{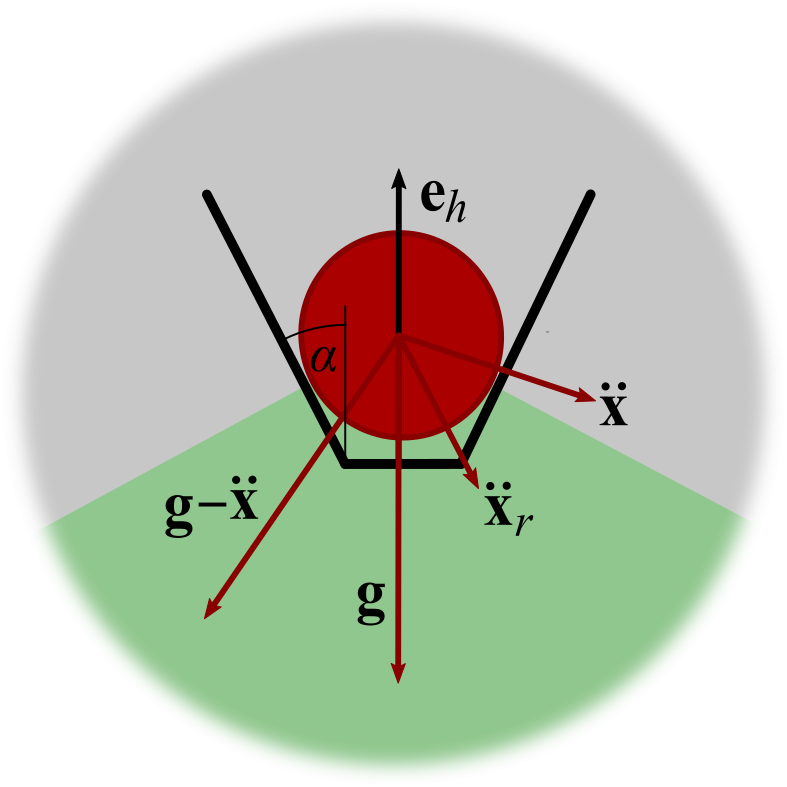}}
    \hspace{0.5cm}
    \caption{(a) In the frictionless case, balls can orbit in the hand. Through sufficient friction, these orbits dissipate, allowing for (b)~the preservation of the ball's resting position, by imposing the constraint $\measuredangle(\handNormal, \gravity - \eeAcc) > \ang{90}+\alpha$, which governs the direction of the gravity-compensated hand acceleration $\eeAcc - \gravity$ based on the hand orientation $\handNormal$ and slope angle $\alpha$.}
    \vspace{-5pt}
\end{figure}

%
Most juggling patterns necessitate distinct touchdown and takeoff positions to prevent collisions between incoming and outgoing balls,
thus requiring a horizontal carry movement during dwell time.
%
When juggling with open, unactuated hands, the ball is not fixed but held in place by gravity and inertia.
Therefore, it may roll out of hand.
This limitation becomes particularly pronounced when initiating a throw from a standstill following a $2$-throw due to a larger downward acceleration required in the initial countermovement.
%
While combinations of cycle and dwell times that prevent balls from rolling out of hand can be found for each throw height in uniform juggling patterns, 
the defined siteswap juggling problem does not allow for throw-height-specific timing variations,
and the problem needs to be addressed explicitly.
%
We therefore introduce the novel \emph{roll-out constraint}
\begin{equation}
     \forall t \in [T_\TD, T_\TO): \measuredangle(\handNormal,\gravity - \eeAcc) > \ang{90}+\alpha,  \label{eq:rollout_constraint}
\end{equation}
%
for cone-shaped hands,
%
restricting hand acceleration such that potential inertia and gravity-induced contact normal forces accelerate the ball into the cone,
as shown in Figure~\ref{fig:rollout}.
%
Note how the downward acceleration of a hand can never exceed gravity unless the hand is turned upside down,
%
and a steeper cone shape with smaller slope angles $\alpha$ are less restrictive on horizontal accelerations.

For this constraint, we assume the quasi-static case of zero relative velocity between hand and ball, 
corresponding to the desired state of the ball remaining at the bottom of the cone throughout the dwell time.
%
In a frictionless scenario, the ball could continuously orbit the perimeter of a static funnel-shaped hand, as illustrated in Figure~\ref{fig:spiral}.
%
This restricts the application of the proposed \emph{roll-out constraint} to cases with sufficient contact friction and damping to dissipate the kinetic energy of orbital relative movements.
%
The case of a $3$-catch-$9$-throw movement is most prone to break this assumption since $3$-balls have the fastest horizontal velocity, and $9$-throws require the largest countermovement.
%

\section{Empirical Evaluation of Pattern Robustness}

To demonstrate the necessity of the novel contact management constraints introduced in Sections~\ref{sec:premature_contacts} and~\ref{sec:contact_management}, and the sufficiency of the complete constraint set,
we compare the proposed method to baseline~\cite{ploeger2022controlling},
using the number of consecutive successful catches as an indicator of robustness.
%
To evaluate the quality of generated trajectories independent of error sources like model uncertainty, signal latency, observation noise, and trajectory tracking control,
we run all experiments in an idealized simulation scenario.
%
Tested siteswap sequences are representative of all possible vanilla siteswap patterns with throw height up to $9$-throws, excluding kinematically infeasible $1$-throws.

\vspace{-0.1cm}
\subsection{Experimental Setup in Mujoco Simulator}

All experiments are performed in a MuJoCo~\cite{todorov2012mujoco} simulation environment,
featuring two 4-DoF Barrett WAM manipulators
with unactuated funnel-shaped hands, as depicted in Figure~\ref{fig:throw_heights}.
Compared to~\cite{ploeger2022controlling}, the funnel is reduced to a diameter of $100$mm at a slope angle of $\alpha=20$deg,
to enable shorter vacant times and less restrictive \emph{roll-out constraints}.
All balls have a diameter of $75$mm, and contacts are modeled with high stiffness and damping,
approximately equivalent to $\SI{100.000}\newton/\mathrm{m}$ $\SI{1.000}\newton\mathrm{s}/\mathrm{m}$.
%
To reduce the impact of tracking controller errors on assessing the planning quality,
we employ an inverse dynamics controller with high-gain PD correction of the reference acceleration.
Applying proportional gain $k_P=2.000$ and derivative gain $k_D=500$ results in sub-millimeter precision at all times.
Directly setting the robot position and velocity at every time step
would lead to artifacts in the contact dynamics,
due to the slight mismatch between the third order trajectory representation and the second order
simulation dynamics.
%
All experiments use the same timing parameters. 
The cycle time $\cycleTime=0.48\si{s}$ trades off between reducing required hand velocities during $9$-throws, which benefit from lower cycle times, and maximizing flight times, which is advantageous for $3$-throws.
The small dwell ratio $r=0.5$ offers extended vacant times required for clean contact switches with unactuated hands.
We consider all siteswap throw heights between $0$ and $9$, excluding $1$-throws,
which would require an actuated wrist for near-vertical take-off
and a dwell ratio $r<0.5$ for non-negative flight times
at constant throw timing.
We discretize each trajectory in 24 steps and apply constraints as shown in Figure~\ref{fig:constraintcycle}.

\vspace{-0.1cm}
\subsection{Stability of Siteswap Patterns and Transitions}

We tested the 98 siteswap patterns listed in Table~\ref{tab:possible_patterns}, which include an exhaustive list of all three to nine-ball siteswaps up to height 9 and length 3.
Considering a pattern empirically stable if it can be continuously juggled for $1000$ consecutive catches.
%
All siteswap patterns are empirically stable when applying the full proposed set of constraints,
%
but only the 10 patterns highlighted in bold font are stable using the baseline without the constraints proposed in Sections~\ref{sec:premature_contacts} and~\ref{sec:contact_management},
clearly highlighting the necessity of these constraints.
The nine-ball cascade pattern can be stabilized using the baseline method
at lower cycle time,
as demonstrated in~\cite{ploeger2022controlling}.

\begin{table}[b]
    \vspace{-0.45cm}
    \centering
    \caption{The tested patterns include all siteswap sequences up to length 3 and throw heights up to 9 excluding $1$-throws. All patterns were executed successfully for $1000$ catches. Bold patterns are also stable when employing the baseline.}
    \vspace{-0.1cm}
    \resizebox{\columnwidth}{!}{%
        \begin{tabular}{|c|c|c|c|c|c|c|c|}
            \hline
            \textbf{3 Balls} & \textbf{4 Balls} & \textbf{5 Balls} & \textbf{6 Balls} & \textbf{7 Balls} & \textbf{8 Balls} & \textbf{9 Balls} \\
            \hline
            \textbf{3}    & 4    & \textbf{5}   & 6    & \textbf{7}    & 8     &  9  \\
            42   & 53   & \textbf{64}  & 75   & \textbf{86}   & 97    &     \\
            423  & 62   & 73  & 84   & 95   & 978   &     \\
            504  & 80   & \textbf{82}  & 93   & \textbf{867}  & 996   &     \\ 
            522  & 534  & 645 & 756  & \textbf{885}  & 9995  &     \\
            603  & 552  & 663 & 774  & 948  & 9968  &     \\
            630  & 633  & 726 & 783  & 966  & 99697 &     \\
            720  & 642  & 744 & 837  & 975  & 99994 &     \\
            900  & 660  & 753 & 855  & 993  &       &     \\
            5304 & 723  & \textbf{807} & 864  & 8884 &       &     \\
            5340 & 750  & 825 & 882  & 9388 &       &     \\
            5520 & 804  & 834 & 936  & 9568 &       &     \\
            6024 & 822  & \textbf{852} & 945  & 9685 &       &     \\
            6330 & 903  & 906 & 963  & 9748 &       &     \\
            7023 & 930  & 933 & 972  & 9784 &       &     \\
            7302 & 5524 & 942 & 990  & 9955 &       &     \\
            8040 & 6055 & 960 & 7773 &      &       &     \\
            9300 & 7333 &     &      &      &       &     \\
            \hline
        \end{tabular}
    }
    \label{tab:possible_patterns}
\end{table}

For the set of tested siteswap patterns, $203$ pairs of patterns with non-trivial transitions exist. 
Continuously navigating back and forth between the corresponding disjoint loops in the siteswap graph,
also succeeded for $1000$ consecutive catches for each pair of patterns.
Video documentation can be found at \href{https://kai-ploeger.com/beyond-cascades}{https://kai-ploeger.com/beyond-cascades}.

\vspace{-0.1cm}
\subsection{Stability of Random Walks on the Siteswap Graph}
\vspace{-0.05cm}

We continuously sample target heights uniformly from all available options for each throw,
performing a random walk on the five-ball siteswap graph.
%
An infinite juggling sequence of this kind includes every finite five-ball juggling sequence.
%
In practice, we perform a random walk for 1.000.000 catches on the strongly connected five-ball siteswap graph of height 9,
Given that the state of all five balls is defined up to throwing accuracy by the previous 9 throws
it is safe to assume that every possible combination of ball states have been visited many times,
demonstrating sufficiency of the proposed constraint set for five-ball siteswap juggling.

\begin{figure}
    \centering   
    \includegraphics[width=\columnwidth, trim={0cm, 0cm , 0cm, 0cm}]{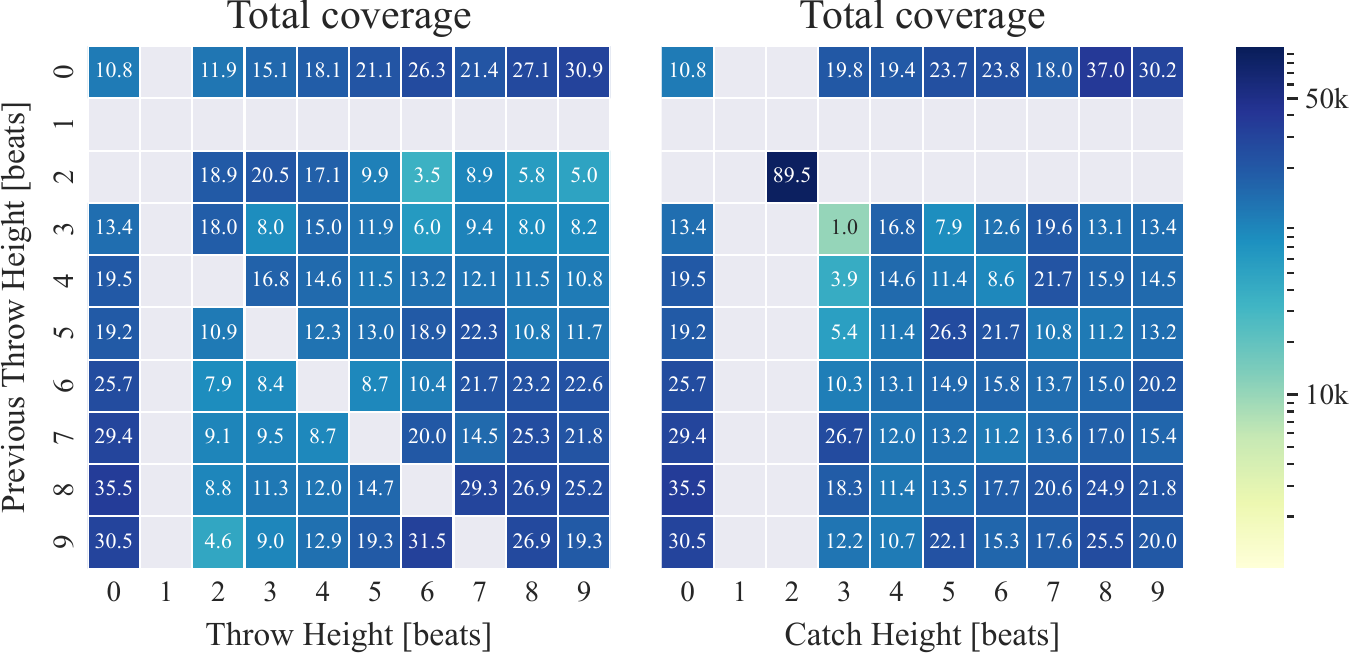}
    \caption{The shape of the planned trajectory primarily depends on the heights of the previous throw, the incoming ball, and the target throw. 
    During a 5-ball random sequence totaling 1.000.000 catches, all possible combinations were frequently encountered.
    An $N$-throw can never be followed by an $N-1$-throw,
    a 2-catch always follows a 2-throw,
    and 1-throws are mechanically infeasible on the investigated system.}
    \label{fig:coverage}
    \vspace{-0.5cm}
\end{figure}

Each catch-and-throw trajectory solely depends on the preceding throw, the target throw height, and the predicted touchdown of the incoming ball.
Planning is independent of all other balls.
Figure~\ref{fig:coverage} illustrates the recorded random walk covering all possible combinations of throw heights for these three relevant throws.
%
Note that an $N$-throw can never be followed by an $N-2$-throw from the same hand
to avoid simultaneous touchdown,
a previous 2-throw requires a subsequent 2-catch as the ball stays in the same hand,
and $1$-throws are kinematically infeasible given the test setup.
%
From the full coverage of all possible trajectory solver inputs without a single dropped ball,
we can conclude that the proposed set of constraints is sufficient to stabilize all possible
siteswap patterns with any number of balls up to throw height 9.
%
In contrast, random walks fail on average after $57$ catches when removing the \emph{roll-out constraint}
and after $7$ catches when removing the \emph{premature contact avoidance constraint},
averaging over $500$ random seeds each.

\section{CONCLUSION}

We introduced novel task-space hand trajectory constraints
for continuous contact management
preventing balls from rolling out of hand during dwell time
and preventing unintended hand-ball contacts during the vacant time.
Through extensive evaluation on various siteswap patterns, pattern transitions, and a randomized sequence of $1.000.000$ throws
in an idealized simulation scenario,
we demonstrate the necessity of the proposed constraints for all possible vanilla siteswap patterns
utilizing throw heights 2-9 and 0,
when extending the base planner introduced in~\cite{ploeger2022controlling}.

The approach is limited through the employed high-level ball trajectory generator,
which does not provide the flexible timing of contact switches required for $1$-throws,
and does not consider possible inter-ball collisions.
Further, the quasi-static assumption of the proposed \emph{roll-out constraint} 
and refrain from touchdown velocity matching
requires sufficient dissipation of kinetic energy in hand-ball contact dynamics.

\subsection{Discussion of Experimental Design Choices}

Deviating from the described experimental setup reveals several insights worth highlighting explicitly:

\vspace{0.1cm}
\subsubsection{Hand Design Choices}
In choosing a hand shape that enables reliable juggling at mixed throw heights,
three main shape properties turned out to be relevant.
Steeper sidewalls holding a carried ball in place relax the \emph{roll-out constraint}, allowing for larger lateral acceleration during the carry phase.
Hands with larger diameters require less accurate ball state estimation and trajectory tracking and can lead to open-loop stable juggling at sufficient throw accuracy.
A lower outer rim reduces the time required for the ball to move in and out of the hand, 
enabling shorter activation of the \emph{post-takeoff} and \emph{pre-touchdown constraints},
and thereby leaving a larger fraction of the vacant time open for less jerky lateral movement.
Since these geometric properties conflict with each other, a careful trade-off needs to be chosen.
If the friction between the surface materials of hand and ball does not dissipate sufficient kinetic energy for the quasi-static assumption of the \emph{roll-out constraint} to hold,
a pyramidal hand shape can prevent the ball from spiraling out of the hand
through repeated damped impacts.

\subsubsection{Throw Accuracy Determined by Physics Parameters}
\begin{figure}[t]
    \centering   
    \includegraphics[width=\columnwidth, trim={0cm, 0cm , 0cm, 0cm}, clip]{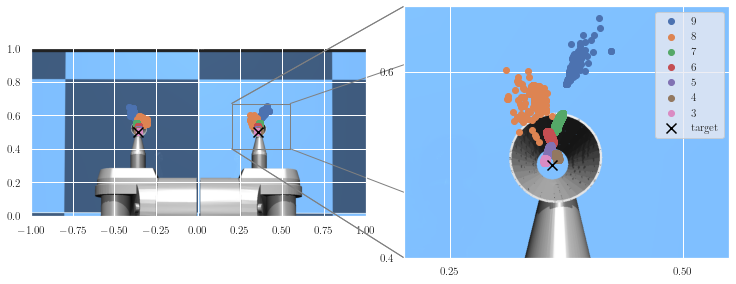}
    \caption{During a random five-ball siteswap sequence, unmodeled contact dynamics introduce slight deviations from the touchdown target, increasing with throw height.}
    \label{fig:throw_accuracy}
    \vspace{-15pt}
\end{figure}

As shown in Figure~\ref{fig:throw_accuracy}, we encounter errors in touchdown positions,
increasing with throw height.
With simultaneously increasing contact stiffness, controller gains, and control frequency at zero friction,
touchdown errors asymptotically approach zero,
indicating that imperfect throw accuracy stems from unmodeled dynamics,
and is not intrinsic to the planned reference trajectory.
While these changes to the environment improve throw accuracy,
they reduce dissipation of kinetic energy,
prohibiting successful catches.

\subsubsection{Choice of Solver Time Discretization}
A finer time discretization of $96$ steps with equivalent constraint activation durations did not impact throw accuracy or stability,
resulting in a significant compute overhead that has no benefit.

\subsection{Real-Time and Real-World Applicability}
On a single CPU core, one catch-and-throw cycle can reliably be planned within 15ms,
which is sufficient for MPC-style online replanning in a real-time setting.
Trajectory tracking errors quickly deteriorate throwing accuracy, especially for high throws.
The presented empirical assessment of planned trajectories relies on sub-millimeter precision trajectory tracking
at peak end effector velocities of $10\si{\meter}/\mathrm{s}$ and peak accelerations of $400\si{\meter}/\mathrm{s}^2$,
which is challenging to achieve on a physical system.

\vspace{-0.15cm}
\subsection{Directions for Future Work}

One evident continuation of this work is application to a physical robot.
This could be in the direct application of the proposed planner to a physical system,
or through utilization of the identified required trajectory properties as inductive biases in a learning scenario.
Furthermore, addressing inter-ball collision avoidance in the ball trajectory generation, 
through takeoff location and timing variation
presents a challenging long-horizon planning problem over multiple future throws.

\section*{ACKNOWLEDGMENT}
This work was supported by the Hessian Ministry of Higher Education, Research, Science and the Arts
and its LOEWE research priority program ‘WhiteBox’ under grant LOEWE/2/13/519/03/06.001(0010)/77.

Grammarly was employed to enhance the readability of this work.


\pagebreak

\enlargethispage{-40\baselineskip}
\bibliographystyle{ieeeconf}
\bibliography{references}

\end{document}